%% file: main.tex
\documentclass[acmsmall, natbib=false]{acmart}
\renewcommand\footnotetextcopyrightpermission[1]{} 
\usepackage{xspace}
\usepackage{enumitem}
\usepackage{xcolor}
\usepackage{array}
\usepackage{float}
\usepackage{url}
\usepackage{graphicx}
\usepackage{booktabs}
\usepackage{multirow}
\usepackage{tabularx}
\usepackage{caption}
\usepackage{geometry}
\usepackage{cite}
\usepackage{hyperref}

\usepackage[most]{tcolorbox}

\tcbuselibrary{skins} 

\tcbset{
    mycardstyle/.style={
        colframe=teal!75,       
        colback=yellow!10,            
        arc=2mm,                      
        boxrule=0.5mm,                
        enhanced,                     
        before=\par\smallskip\noindent, 
        leftrule=1mm,                 
        rightrule=0.5mm,
        toprule=0.5mm,
        bottomrule=0.5mm
    }
}

\begin{document}

\title{``Spliced, or not spliced, that is the question'': Can ChatGPT Perform Image Splicing Detection? A Preliminary Study}

\author{Souradip Nath}
\email{snath8@asu.edu}
\orcid{0000-0001-7228-2316}
\affiliation{
    \institution{Arizona State University}
  \city{Tempe}
  \state{Arizona}
  \country{USA}
}


\begin{abstract}
Multimodal Large Language Models (MLLMs) like GPT-4V are capable of reasoning across text and image modalities, showing promise in a variety of complex vision-language tasks. In this preliminary study, we investigate the out-of-the-box capabilities of GPT-4V in the domain of image forensics, specifically, in detecting image splicing manipulations. Without any task-specific fine-tuning, we evaluate GPT-4V using three prompting strategies: Zero-Shot (ZS), Few-Shot (FS), and Chain-of-Thought (CoT), applied over a curated subset of the CASIA v2.0 splicing dataset. 

Our results show that GPT-4V achieves competitive detection performance in zero-shot settings (more than 85\% accuracy), with CoT prompting yielding the most balanced trade-off across authentic and spliced images. Qualitative analysis further reveals that the model not only detects low-level visual artifacts but also draws upon real-world contextual knowledge such as object scale, semantic consistency, and architectural facts, to identify implausible composites. While GPT-4V lags behind specialized state-of-the-art splicing detection models, its generalizability, interpretability, and encyclopedic reasoning highlight its potential as a flexible tool in image forensics.
\end{abstract}





\maketitle

\input{1_introduction}
\input{3_methodology}
\input{4_results}
\input{5_analysis}

\input{6_limitations}
\input{7_conclusion}

\bibliographystyle{acm}
\bibliography{references}

\end{document}

%% file: 1_introduction.tex
\section{Introduction}
\label{sec:intro}

With the rapid advancement of digital image processing technologies and the widespread availability of image editing tools, image manipulation has emerged as a critical concern in the field of image forensics. Among various manipulation techniques, image splicing~\cite{nath2021automated} and copy-move forgery~\cite{al2013passive} are the two most prevalent forms of tampering. 

In a copy-move forgery, content from one region of an image is duplicated and pasted into another region within the same image. In contrast, image splicing (or compositing) involves inserting content from an external source into an authentic image, with the intent of creating a visually seamless composite that appears natural and unaltered. Due to sophisticated post-processing operations—such as compression, geometric deformation, edge softening, blurring, and smoothing—detecting such manipulations with the naked eye becomes exceedingly difficult. As a result, the image forensics community has devoted considerable effort to developing automated techniques capable of accurately identifying spliced images.

Traditionally, image splicing detection relied heavily on hand-crafted features derived from the spectral and visual information embedded within images. However, in recent years, deep neural networks such as Deep Autoencoders~\cite{zhou2017anomaly}, Convolutional Neural Networks (CNNs)\cite{o2015introduction}, and advanced variants such as Mask R-CNNs\cite{he2017mask} have demonstrated remarkable capabilities in automatically extracting complex features from high-dimensional data. These models generalize effectively across various computer vision tasks~\cite{rao2016deep}. As a result, significant research efforts have focused on leveraging these deep learning techniques for automated image forgery detection and localization~\cite{nath2021automated,meena2021deep,tallapragada2024blind,bhowal2024deep,hosny2023new,kasim2024deep,mallick2022copy}. While effective, these models often struggle with generalization across domains and offer limited transparency in their decision-making processes.

More recently, Large Language Models (LLMs) have revolutionized the field of natural language processing (NLP). While LLMs have demonstrated impressive reasoning capabilities across a wide range of NLP tasks, they are inherently "blind" to visual information, as they operate solely on discrete text inputs. However, with the release of GPT-4V~\cite{achiam2023gpt}, there has been a surge of interest in Multimodal LLMs (MLLMs). MLLMs integrate powerful LLMs with pretrained modality-specific encoders—such as vision or audio encoders—connected through learnable interfaces. 
This architecture enables the model to align and process multimodal inputs, allowing for capabilities like image captioning, visual question answering, and OCR-free reasoning.
Unlike earlier multimodal models that were constrained by task-specific designs, MLLMs benefit from massive parameter scales and novel training paradigms like multimodal instruction tuning. 

While the full potential of MLLMs across domains remains an active area of research, this work offers a preliminary investigation into the capabilities of MLLMs—specifically GPT-4—for the task of automated image splicing detection.
Our objective is not to outperform specialized detectors, but to understand whether a general-purpose MLLM can reason about image authenticity out of the box. We employ three prompting strategies, Zero-Shot~\cite{radford2019language}, Few-Shot~\cite{brown2020language}, and Chain-of-Thought~\cite{wei2022chain} and analyze the model’s performance across a curated subset of the CASIA v2.0 dataset~\cite{casia}. Our findings indicate that GPT-4V demonstrates surprisingly strong baseline performance, particularly in zero-shot settings, and that it can detect tampering not just through visual artifacts but also by leveraging world knowledge about object relationships, scale, and semantics.

Through quantitative evaluation and qualitative insights, we uncover the model’s strengths, limitations, and potential as a complementary tool for image forensics. The code repository for this work can be found at~\cite{github}.

%% file: 3_methodology.tex
\section{Experimental Setup}
\label{sec:methodology}

\begin{figure}
    \centering
    \includegraphics[width=\linewidth]{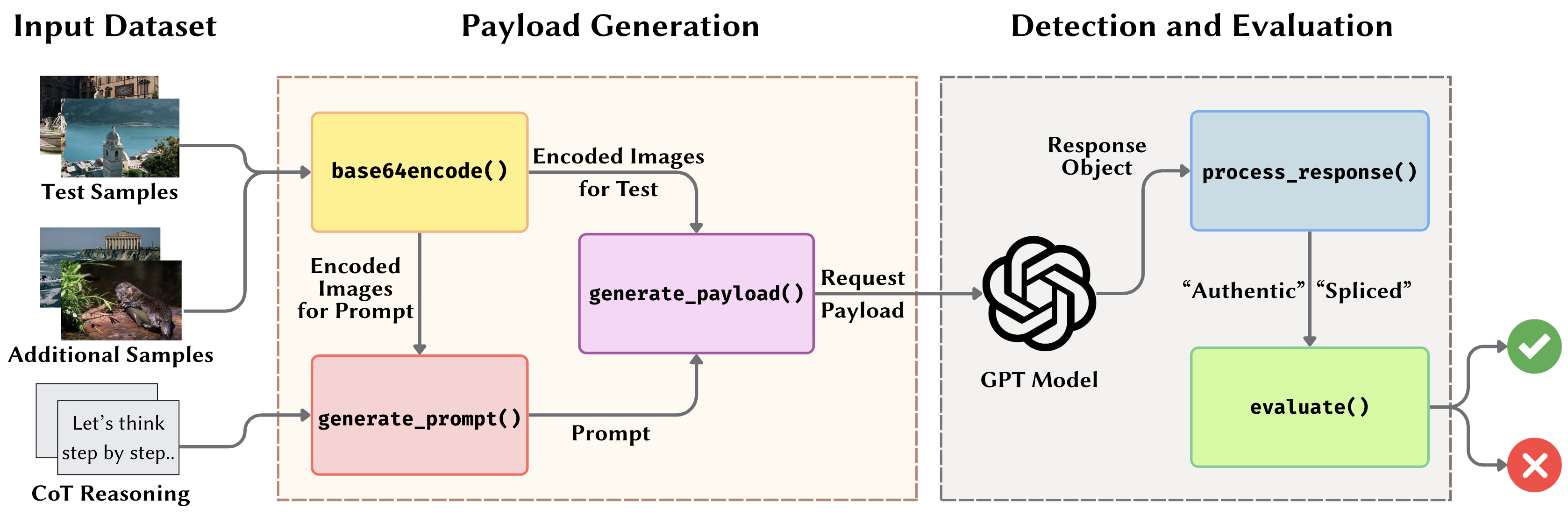}
    \caption{\textbf{Overview of the Experimental Framework.} Test samples and additional prompt samples are first encoded to Base64 format. Prompts, optionally including example images and Chain-of-Thought reasoning depending on the prompt strategy, are constructed and combined with the encoded images to generate a request payload for the GPT-4.1 model. The model's responses are processed and evaluated to determine the detection accuracy.}
    \label{fig:methodology}
\end{figure}

To evaluate the capabilities of a Multimodal Large Language Model (MLLM) in the task of image splicing detection, we design a comprehensive experimental framework, as illustrated in Figure~\ref{fig:methodology}. This section details the dataset used, the prompting strategies employed, the model configuration, and the evaluation metrics adopted in our study.

\subsection{Dataset}

In this work, we conduct our experiments using the CASIA v2.0 dataset~\cite{casia}, a widely adopted benchmark for research in image splicing detection and localization. The dataset comprises a total of 12,614 labeled RGB images, including 7,491 authentic (Au) and 5,123 tampered (Tp) images. The tampered set encompasses both image splicing (1,849 images) and copy-move forgery (3,274 images), as the dataset was originally curated to support research on both manipulation techniques.

\begin{figure}[t]
    \centering
    \includegraphics[width=\linewidth]{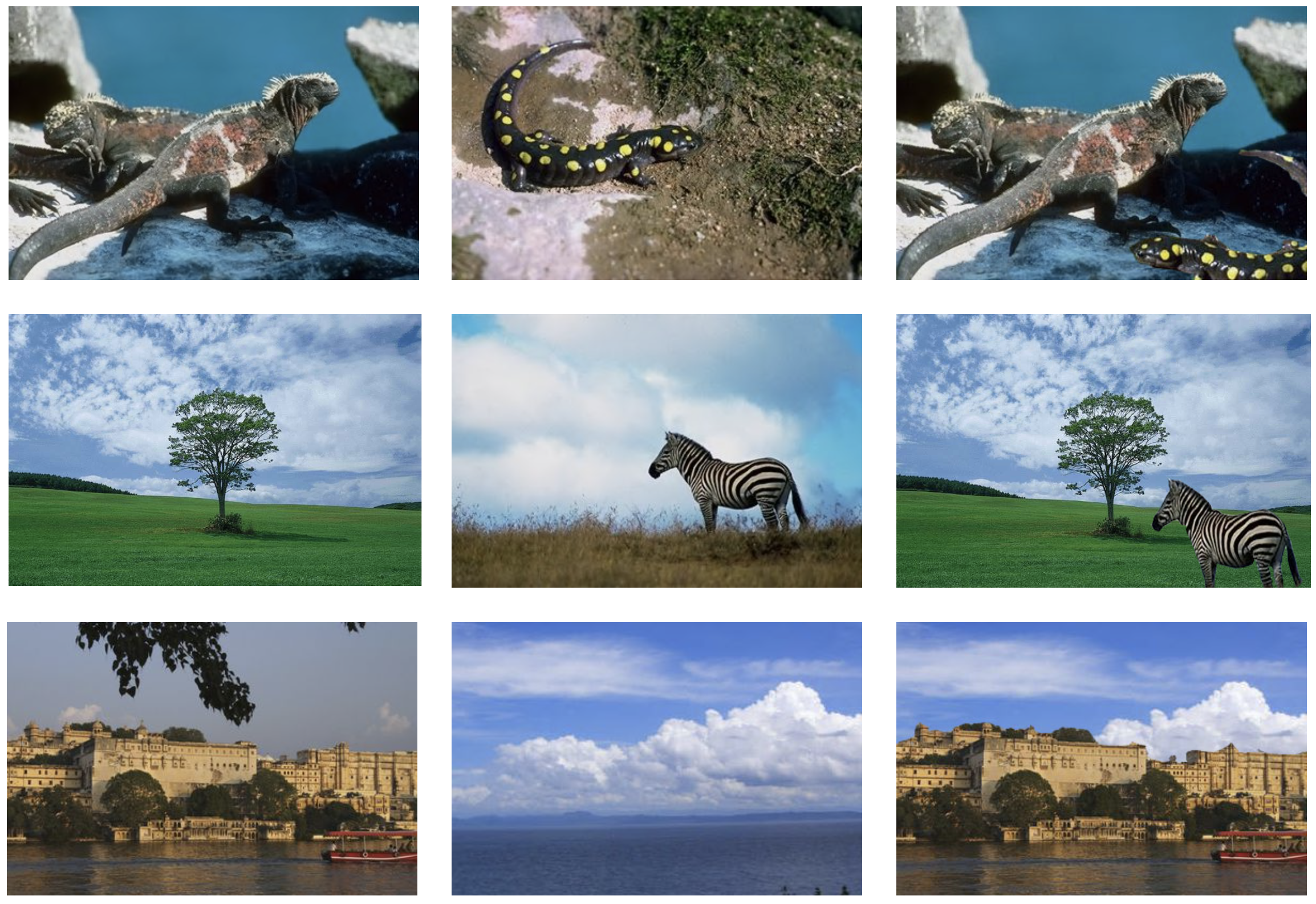}
    \caption{\textbf{Examples of Image Splicing from the CASIA v2.0 Dataset.} In each row, the first image is the source (base) image, and the second is the target image from which a region is extracted and inserted into the source. The resulting spliced image is shown third. Row 1 illustrates a same-category splicing example, while Rows 2 and 3 demonstrate cross-category splicing.}
    \label{fig:casia}
\end{figure}

The authentic images span a diverse array of categories, including animals, characters, architecture, nature, and indoor scenes. For splicing, pairs of authentic images, either from the same or different categories, are used to generate manipulated composites. Figure~\ref{fig:casia} illustrates representative examples of authentic and spliced images sampled from the dataset. These spliced images are further subjected to various post-processing operations, including resizing, deformation, and blurring, to enhance visual realism and better simulate real-world forgeries.

Given the scope of this work and the need for a streamlined experimental design, we curated a subset of the dataset following these steps:

\begin{itemize}[leftmargin=*]
    \item We first excluded all copy-move forgery images, focusing exclusively on image splicing cases.
    \item Next, we discarded uncompressed image formats (e.g., TIFF), retaining only JPEG images for consistency and efficient processing.
    \item We then filtered the dataset to include only the three most frequent categories—animal, architecture, and character—yielding 3,138 authentic images. For spliced images, we retained only those where either the source or target belonged to one of these categories, resulting in a total of 650 spliced images.
    \item Finally, we randomly sampled 3,100 authentic and 600 spliced images for use in our experiments. The remaining images were reserved for use in constructing few-shot and Chain-of-Thought prompts, as described in Section~\ref{subsec:prompt}.
\end{itemize}

The curated dataset used for our evaluation, along with the additional images employed in few-shot and Chain-of-Thought prompts, is available in our project repository~\cite{github}.

\subsection{Prompt Strategy}
\label{subsec:prompt}
Given that the core objective of this work is to evaluate whether an MLLM can perform image splicing detection without any additional training or fine-tuning, we employ the following prompt strategies in our experiments.\smallskip

\noindent
\textbf{Zero-Shot (ZS) Prompting:} As introduced by Radford et al.~\cite{radford2019language}, zero-shot prompting involves providing the model with a task description without supplying any labeled examples during inference. The model relies entirely on its pre-trained knowledge to perform the task, thus eliminating the need for extensive training data or task-specific fine-tuning.

For the image splicing detection task, we use the following prompt:
\begin{tcolorbox}[mycardstyle]
    \emph{Inspect the provided image and identify whether it is original or has been spliced. Answer with `Authentic' for an unedited image, or `Spliced' for a manipulated one.}
\end{tcolorbox}

\noindent   
\textbf{Few-Shot (FS) Prompting:} In this approach, the model is provided with a small number of input-output examples to guide its understanding of the task~\cite{brown2020language}. For each image splicing detection query, we supply the model with four illustrative examples, two authentic and two spliced.

To improve contextual relevance, we apply category matching when selecting examples. For instance, if the input image belongs to the `animal' category and is authentic, we include authentic examples from the same category, as well as spliced examples in which either the source or target involves an `animal' image. This can help the model reason based on visual and semantic consistency within similar content types.

The final prompt structure is as follows:
\begin{tcolorbox}[mycardstyle]
\emph{Inspect the provided image and identify whether it is original or has been spliced. Answer with `Authentic' for an unedited image, or `Spliced' for a manipulated one.\\
For example, this image is {[Authentic or Spliced]}.\\
{[Attached image here]}\
{[Repeated for four example images]}
}
\end{tcolorbox}

\noindent 
\textbf{Chain-of-Thought (CoT) Prompting:} Chain-of-Thought prompting~\cite{wei2022chain} is a technique designed to elicit more structured and deliberate reasoning from LLMs by encouraging step-by-step explanations rather than direct answers. This approach has been shown to improve performance, particularly on tasks that require multi-step inference~\cite{sahoo2024systematic}. In our work, we adopt CoT prompting to guide the model through the reasoning process and to mitigate potential biases introduced by the examples used in the few-shot prompting strategy.

\begin{figure}[t]
    \centering
    \includegraphics[width=1\linewidth]{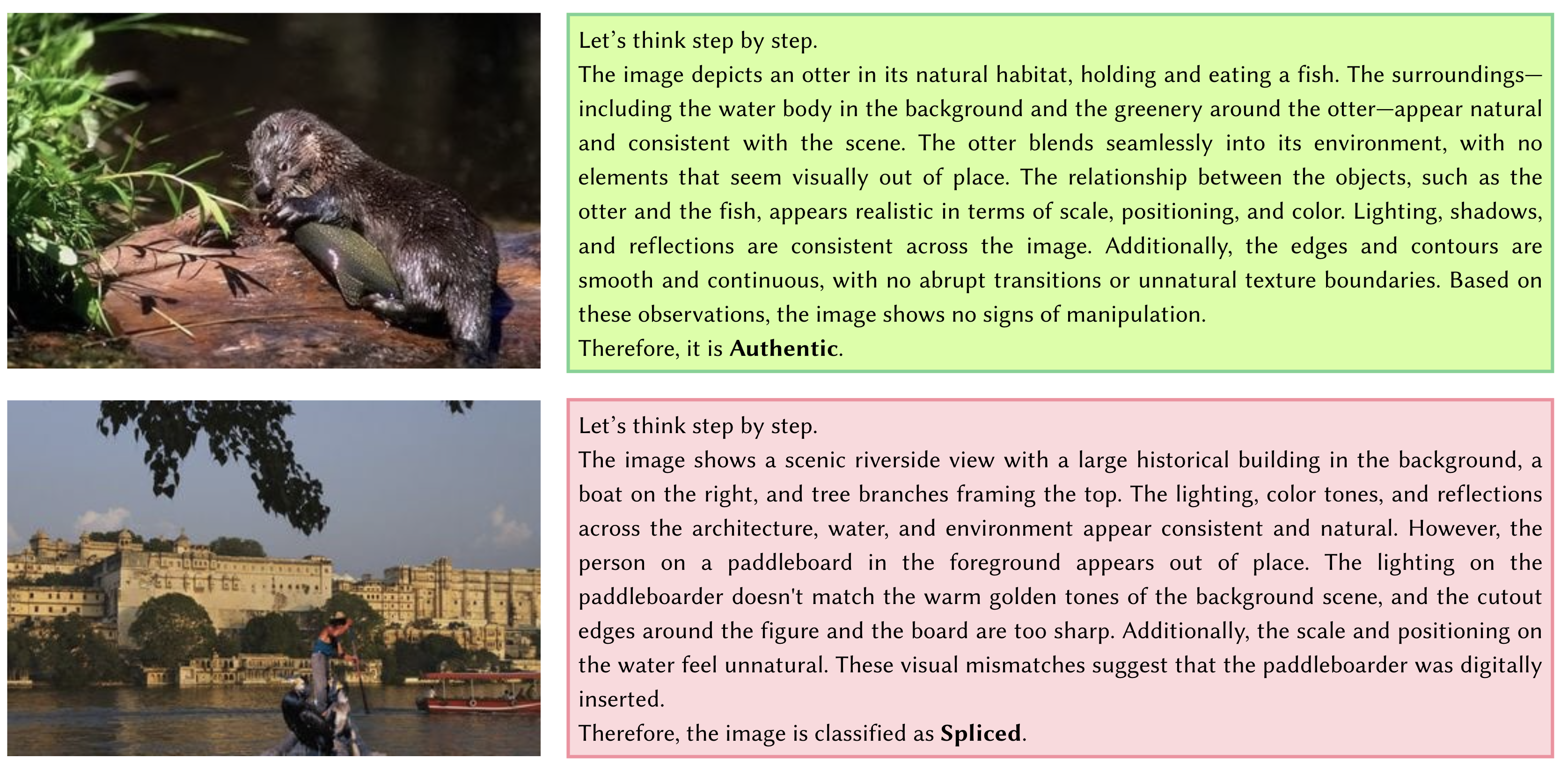}
    \caption{\textbf{Examples of Chain-of-Thought (CoT) Prompts.} Each prompt includes an image along with a detailed, step-by-step explanation guiding the model's decision. }
    \label{fig:cot-example}
\end{figure}

For each detection query, we provide the model with four annotated examples, two authentic and two spliced, accompanied by manually crafted step-by-step explanations of the detection rationale. These explanations help the model contextualize and justify the decision-making process. Figure~\ref{fig:cot-example} presents examples of CoT prompts for both an authentic and a spliced image.

The final prompt structure is as follows:

\begin{tcolorbox}[mycardstyle]
\textit{Inspect the provided image and identify whether it is original or has been spliced. Answer with `Authentic' for an unedited image, or `Spliced' for a manipulated one.\\
For example, this image is {[Authentic or Spliced]}.\\
{[Attached reasoning here]}\
{[Attached image here]}\
{[Repeated for four example images]}
}
\end{tcolorbox}

\subsection{Model}

We utilize the GPT-4.1 model (version: 2025-04-14) for evaluation in this study, as it is currently recognized as the most capable publicly available multimodal model for complex reasoning tasks, according to OpenAI’s documentation\footnote{\url{https://platform.openai.com/docs/models/gpt-4.1}}. To ensure the reproducibility of our results, we fix the temperature parameter to 0, making the model's outputs deterministic for identical inputs.

To interface with the GPT-4.1 model via the OpenAI API, we encode each input image as a Base64 string. This transformation converts the binary image file into a textual format suitable for transmission over the API, as required by the model’s input specifications. The Base64-encoded string is then embedded in the request payload to enable visual input processing by the model.

\subsection{Evaluation Metric}
\label{subsec:metric}

In our experiments, we use \textbf{detection accuracy} as the primary metric to evaluate the performance of the GPT model on the image splicing detection task. Accuracy is defined as the ratio of correctly classified samples (sum of true positives and true negatives) to the total number of samples, encompassing both correct and incorrect predictions.

Given the class imbalance in the dataset with differing numbers of authentic and spliced images, we also compute class-wise accuracy to assess the model's performance on each class independently. Additionally, to investigate whether the semantic category of the image affects model performance, we also observe accuracy scores across individual image categories.

%% file: 4_results.tex
\section{Results}
\label{sec:results}

\input{tables/results}

Table~\ref{tab:results} presents the experimental results from our evaluation. As outlined in Section~\ref{subsec:metric}, we report three primary metrics for each of the prompting strategies: the number of correct detections, incorrect detections, and overall detection accuracy. In addition to the aggregate results, we provide a breakdown of performance across individual classes (authentic vs. spliced) and across image categories (animal, architecture, and character).

For the following analysis, we consider “spliced” as the positive class and “authentic” as the negative class. Based on our evaluation, we highlight the following key findings:

\smallskip\noindent
\textbf{Zero Shot prompting (ZS) performs relatively well:} Overall, ZS achieves strong baseline performance in detecting both authentic and spliced images, with detection accuracy exceeding 85\%. The results suggest that GPT-4V's intrinsic understanding of image splicing is sufficiently robust in zero-shot settings. This effectiveness may stem from the clarity and focus of the direct task instruction, which guides the model without the influence of potentially biased examples. However, the model exhibits a notable false positive rate—misclassifying 370 authentic images as spliced (approximately 12\%)—indicating that even in the absence of examples, the prompt alone can introduce bias. This reflects a possible over-sensitivity to manipulation cues, leading the model to incorrectly flag unaltered images as tampered.

\smallskip\noindent
\textbf{Few-shot prompting (FS) introduces a notable bias:}
Comparing the outcomes of ZS and FS prompting reveals that FS induces a notable bias toward predicting images as authentic, an exact opposite behavior from ZS. This is evidenced by an increase in both true negatives—from 2,730 to 3,017 (an 11\% gain)—and false negatives—from 87 to 141 (a 62\% increase). Consistent with prior work~\cite{sahoo2024systematic}, this shift underscores how the prompting strategy and the selection of examples can strongly influence the model's decision boundary.

The consequence of this bias is twofold. On one hand, FS improves the model’s confidence and accuracy in identifying authentic images, likely due to the contextual guidance provided by in-domain examples. This leads to a substantial 9\% improvement in detection accuracy for authentic cases. On the other hand, this gain comes at the cost of reduced performance in detecting spliced images, as the model increasingly misclassifies them as authentic. As prior work has observed~\cite{sahoo2024systematic}, when few-shot examples lack diversity or fail to represent subtle cues, the model tends to overgeneralize, introducing biases in response.

\smallskip\noindent
\textbf{Chain-of-Thought improves FS-induced bias:}
Analysis of the results from CoT prompting reveals that it effectively mitigates the bias introduced by the FS strategy, primarily by reducing the overall number of authentic classifications. As a consequence, CoT improves spliced image detection across all three categories, resulting in an overall accuracy gain of approximately 5\% compared to FS. This improvement is likely due to the step-by-step reasoning process, which enables the model to better identify subtle inconsistencies within images—an important signal for detecting manipulations.

Although the CoT strategy slightly reduces detection accuracy for authentic images, possibly due to overexplaining simple visual content, which may introduce unnecessary ambiguity, it still outperforms ZS prompting by a significant margin. Overall, CoT offers a strong middle ground, delivering balanced performance across both authentic and spliced classes while leveraging its reasoning capabilities when appropriate.\smallskip

\noindent
\textbf{Performance Variation across categories is different for authentic vs. spliced:} For authentic image detection, the model demonstrates consistent performance across all three prompting strategies, with minimal variation between categories. This suggests that identifying unaltered content is relatively stable and less influenced by the semantic category of the image.

In contrast, the detection of spliced images reveals significant performance variation across categories. Notably, the model struggles the most with spliced images in the Architecture category across all prompting strategies—particularly under FS prompting. This could be attributed to the complex textures, repetitive patterns, and structured layouts common in architectural imagery, which may obscure splicing artifacts. Moreover, the introduction of few-shot examples appears to reinforce the model’s bias toward authentic classification in this category, leading to a substantial rise in false negatives (from 17 to 34).

Conversely, the Animal category exhibits the most consistent and high detection performance for spliced images, possibly due to the irregular shapes, organic textures, and natural backgrounds in animal images, which make manipulations more visually distinct and easier for the model to detect. 

%% file: tables/results.tex
\begin{table*}[]
\centering
\resizebox{\textwidth}{!}{%
\begin{tabular}{@{}lccccccccc@{}}
\toprule
          &              \multicolumn{3}{c}{\textbf{Correct Detection}} & \multicolumn{3}{c}{\textbf{Incorrect Detection}} & \multicolumn{3}{c}{\textbf{Accuracy}} \\ 
          &               ZS     & FS     & FS-CoT    & ZS      & FS     & FS-CoT     & ZS     & FS     & FS-CoT     \\ \midrule
Authentic               & \textbf{2,730}      & \textbf{3,017}      & \textbf{2,981}         & \textbf{370}       & \textbf{83}      & \textbf{119}          & \textbf{88.07\%}     & \textbf{97.32\%}     & \textbf{96.16\%}          \\
           \hspace{0.5cm}Animal       & 940      & 1,042      & 1,041         &  141       &  39     & 40          & 86.96\%      & 96.39\%      & 96.30\%          \\
          \hspace{0.5cm}Architecture & 912      & 1,028      & 1,012         & 139       & 23      & 39          & 86.77\%      & 97.81\%      & 96.29\%          \\
           \hspace{0.5cm}Character    & 878      & 947      & 928         & 90       & 21     & 40          & 90.70\%      & 97.83\%      & 95.87\%           \\ \midrule
Spliced                 & \textbf{513}      & \textbf{459}      & \textbf{488}         & \textbf{87}       & \textbf{141}      & \textbf{112}          & \textbf{85.50\%}      & \textbf{76.50\%}      & \textbf{81.33\%}          \\
          \hspace{0.5cm}Animal       & 200      & 186      & 196        & 22       & 36      & 26          & 90.09\%      & 83.78\%      & 88.29\%          \\
          \hspace{0.5cm}Architecture & 61      & 44      & 49         & 17       & 34      & 29          & 78.21\%     & 56.41\%      & 62.82\%          \\
          \hspace{0.5cm}Character    & 269      & 245      & 260         & 49       & 73      & 58          & 84.59\%      & 77.04\%      & 81.76\%          \\ \midrule
Overall                 &  \textbf{3,243}     & \textbf{3,476}      & \textbf{3,469}         & \textbf{457}       & \textbf{224}      & \textbf{231}          & \textbf{87.65\%}      & \textbf{94.02\%}      & \textbf{93.76\%}          \\ \bottomrule

\end{tabular}
}
\vspace{5pt}
\caption{\textbf{Experimental Results.} For each of the three prompting strategies, Zero-Shot (ZS), Few-Shot (FS), and Few-Shot Chain-of-Thought (FS-CoT), a number of correct detections, incorrect detections, and the detection accuracy are reported.}
\label{tab:results}
\end{table*}

%% file: 5_analysis.tex
\section{Qualitative Analysis}
\label{sec:discussion}  

In addition to our quantitative evaluation of GPT-4V’s capabilities in image splicing detection, we conducted a qualitative analysis to examine the model’s reasoning process. Notably, in the ZS prompting strategy, the model occasionally supplemented its classification decision with natural language explanations—since we did not constrain it to output only a label. This provided a valuable opportunity to analyze how the model interprets visual evidence and arrives at its conclusions.

Across the dataset, we found that GPT-4V often cited a consistent set of visual cues in its explanations. These include discrepancies in lighting and shadows, edge blending artifacts, texture mismatches, differences in focus or resolution, and inconsistencies in brightness or color tones. This is expected, as these are well-known indicators of manipulation in image forensics, and in the absence of examples or explicit reasoning chains, the model appears to fall back on these learned visual heuristics.

\begin{figure}[t]
    \centering
    \includegraphics[width=1\linewidth]{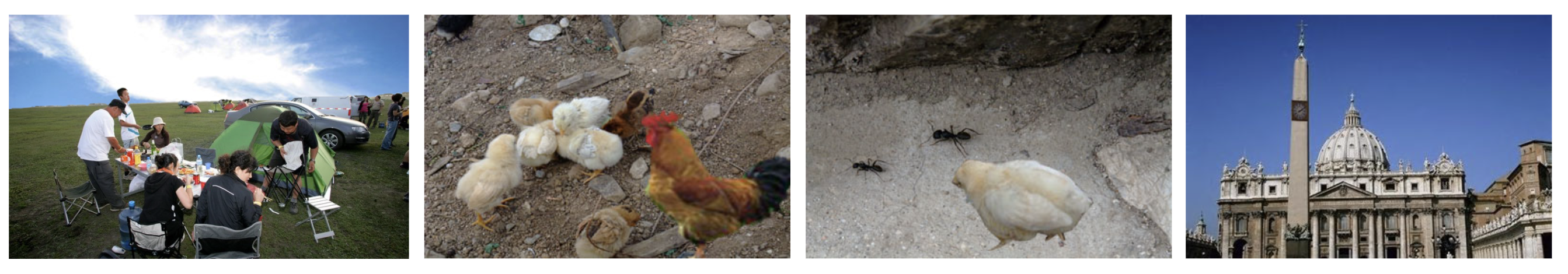}
    \caption{Sample Images for Qualitative Analysis.}
    \label{fig:analysis}
\end{figure}

For instance, for the first image in Figure~\ref{fig:analysis}, the model explains:
\begin{quote}
The \emph{lighting} and \emph{shadows} on the people and objects in the foreground \emph{do not match the brightness and color tone} of the sky in the background, indicating that the image has likely been manipulated. The sky appears \emph{unusually bright and smooth} compared to the rest of the scene, which is a common sign of splicing.
\end{quote}

Similarly, in the second image, it notes:
\begin{quote}
The rooster in the foreground appears to be edited into the image. Its \emph{edges are noticeably sharper}, and the \emph{lighting and focus} do not match the rest of the scene, indicating manipulation.
\end{quote}

However, our analysis also uncovered more sophisticated forms of reasoning that suggest GPT-4V is not limited to surface-level visual artifacts. In several instances, the model drew upon contextual and world knowledge to assess the plausibility of the scene—indicating that its judgments were influenced by an internal understanding of real-world semantics.

In one such example (third image in Figure~\ref{fig:analysis}), the model evaluates the scale of objects:
\begin{quote}
The image appears to be manipulated. \emph{The chick is disproportionately small compared to the ants}, indicating that elements from different images have been combined.
\end{quote}

In another case (spliced image in the first row in Figure~\ref{fig:casia}), the model reasons based on natural habitat expectations:
\begin{quote}
The image shows a group of marine iguanas, but there is also a \emph{salamander-like creature with bright yellow spots} in the foreground \emph{that does not naturally belong in this environment}. This suggests the image has been manipulated or spliced.
\end{quote}

Even more striking is the model’s use of factual knowledge about iconic landmarks, as seen in the final image of Figure~\ref{fig:analysis}:
\begin{quote}
The image shows \emph{an obelisk in front of St. Peter's Basilica with a clock face} that has been digitally added, which is \emph{not present in the original structure}.
\end{quote}

These examples highlight GPT-4V’s unique ability to combine visual pattern recognition with real-world reasoning—something that traditional deep learning models lack. Unlike models that rely purely on visual inconsistencies, GPT-4V draws upon its multimodal and encyclopedic training to detect implausibilities based on relative object sizes, environmental contexts, or even architectural accuracy. This qualitative analysis demonstrates that, under suitable prompting, multimodal large language models can exhibit human-like reasoning about image content, leveraging both perceptual features and common-sense knowledge to detect image manipulation.


%% file: 6_limitations.tex
\section{Limitations and Future Work}
As seen in similar studies evaluating the capabilities of Large Language Models (LLMs)—particularly ChatGPT—for novel tasks such as sentiment analysis~\cite{wang2023chatgpt} and image classification~\cite{wu2025experimental}, our work is subject to comparable limitations. These include, but are not limited to, the following:

\begin{enumerate}[leftmargin=*]

\item\textbf{Data Leakage.} A key challenge in evaluating Large Language Models—especially closed-source ones like GPT-4—is the uncertainty regarding training data exposure. Since models like ChatGPT are trained on vast, undisclosed corpora, we cannot definitively ensure that the images or their associated metadata used in our experiments were not seen during pre-training. This raises concerns about potential data leakage, which could inadvertently inflate performance metrics. Despite this, we observe notable cases of misclassification, suggesting that the model has not fully memorized or mastered the task.

\item \textbf{Prompt Design.} We do not engage in extensive prompt engineering or optimization, instead relied on simple, intuitive instructions and examples. As observed in our results, few-shot prompting strategy introduced biases, affecting the detection accuracy. Therefore, this approach leaves room for performance improvement through more sophisticated prompt crafting. 

\item \textbf{Limited Model Scope:} Our evaluation focuses solely on GPT-4; we do not include comparisons with other strong MLLMs (e.g., Gemini, Claude, or open-source vision-language models). Although GPT-4V represents the current state-of-the-art in multimodal LLMs, future work should aim to benchmark across a broader spectrum of models to validate generalizability and fairness.
\end{enumerate}

%% file: 7_conclusion.tex
\section{Conclusion}
\label{sec:conclusion}

In this work, we explored the capabilities of GPT-4V, a multimodal large language model (MLLM) in detecting image splicing under three prompting strategies: zero-shot (ZS), few-shot (FS), and chain-of-thought (CoT). Our results show that GPT-4V, even without any task-specific training, achieves strong baseline performance, with over 85\% detection accuracy in the ZS setting. While FS prompting improves the detection of authentic images, it introduces bias that reduces sensitivity to spliced content. CoT prompting effectively mitigates this bias, providing more balanced performance by guiding the model through structured reasoning.

Although GPT-4V currently lags behind traditional state-of-the-art models specifically trained for splicing detection in terms of raw detection performance, its out-of-the-box capabilities are noteworthy. The model not only demonstrates reasonable accuracy without retraining but also exhibits an ability to reason using both visual inconsistencies and contextual world knowledge—such as object scale, natural habitats, and even architectural details. This combination of perceptual and encyclopedic knowledge positions GPT-4V as a promising complementary tool in image forensics, especially in scenarios where flexibility, generalizability, and human-like interpretability are valued.